\documentclass[10pt,twocolumn,letterpaper]{article}

\usepackage[pagenumbers]{wacv} 

\newcommand{\colorize}[1]{{\color{blue}#1}}

\newcommand{\ourMethod}{LinDeps\xspace}

\definecolor{wacvblue}{rgb}{0.21,0.49,0.74}
\usepackage[pagebackref,breaklinks,colorlinks,allcolors=wacvblue]{hyperref}

\def\confYear{2026}

\title{LinDeps: A Fine-tuning Free Post-Pruning Method to Remove Layer-Wise Linear Dependencies with Guaranteed Performance Preservation}

\author{Maxim Henry, Adrien Deli\`ege, Anthony Cioppa, and Marc Van Droogenbroeck\\
Montefiore Institute, University of Li\`ege, Li\`ege, Belgium\\
{\tt\small \{Maxim.Henry,Adrien.Deliege,Anthony.Cioppa,M.VanDroogenbroeck\}@uliege.be}
}

\newcommand{\mysection}[1]{\vspace{2pt}\noindent\textbf{#1}}

\newcommand{\batch}{B}            
\newcommand{\channels}{C}         
\newcommand{\channelsprime}{C'}   
\newcommand{\height}{H}           
\newcommand{\width}{W}            
\newcommand{\A}{\mathbf{A}}       
\newcommand{\Aprime}{\mathbf{A}'} 
\newcommand{\R}{\mathbf{R}}       
\newcommand{\Q}{\mathbf{Q}}       
\newcommand{\Pmat}{\mathbf{P}}    
\newcommand{\Lmat}{\mathbf{L}}    
\newcommand{\taup}{\tau}          

\begin{document}
\maketitle

\begin{abstract}
Convolutional Neural Networks (CNN) are widely used in many computer vision tasks. 
Yet, their increasing size and complexity pose significant challenges for efficient deployment on resource-constrained platforms.
Hence, network pruning has emerged as an effective way of reducing the size and computational requirements of neural networks by removing redundant or unimportant parameters.
However, a fundamental challenge with pruning consists in optimally removing redundancies without degrading performance. 
Most existing pruning techniques overlook structural dependencies across feature maps within a layer, resulting in suboptimal pruning decisions.
In this work, we introduce \emph{\ourMethod}, a novel post-pruning method, \ie, a pruning method that can be applied on top of any pruning technique, which systematically identifies and removes redundant filters via linear dependency analysis.
Particularly, \ourMethod applies pivoted QR decomposition to feature maps to detect and prune linearly dependent filters. 
Then, a novel signal recovery mechanism adjusts the next layer’s kernels to preserve compatibility and performance without requiring any fine-tuning.
Our experiments on CIFAR-10 and ImageNet with VGG and ResNet backbones demonstrate that \ourMethod improves compression rates of existing pruning techniques while preserving performances, leading to a new state of the art in CNN pruning. 
We also benchmark \ourMethod in low-resource setups where no retraining can be performed, which shows significant pruning improvements and inference speedups over a state-of-the-art method.
\ourMethod therefore constitutes an essential add-on for any current or future pruning technique.
\end{abstract}

\section{Introduction}\label{sec:introduction}

\begin{figure}
    \centering
    \includegraphics[width=\linewidth]{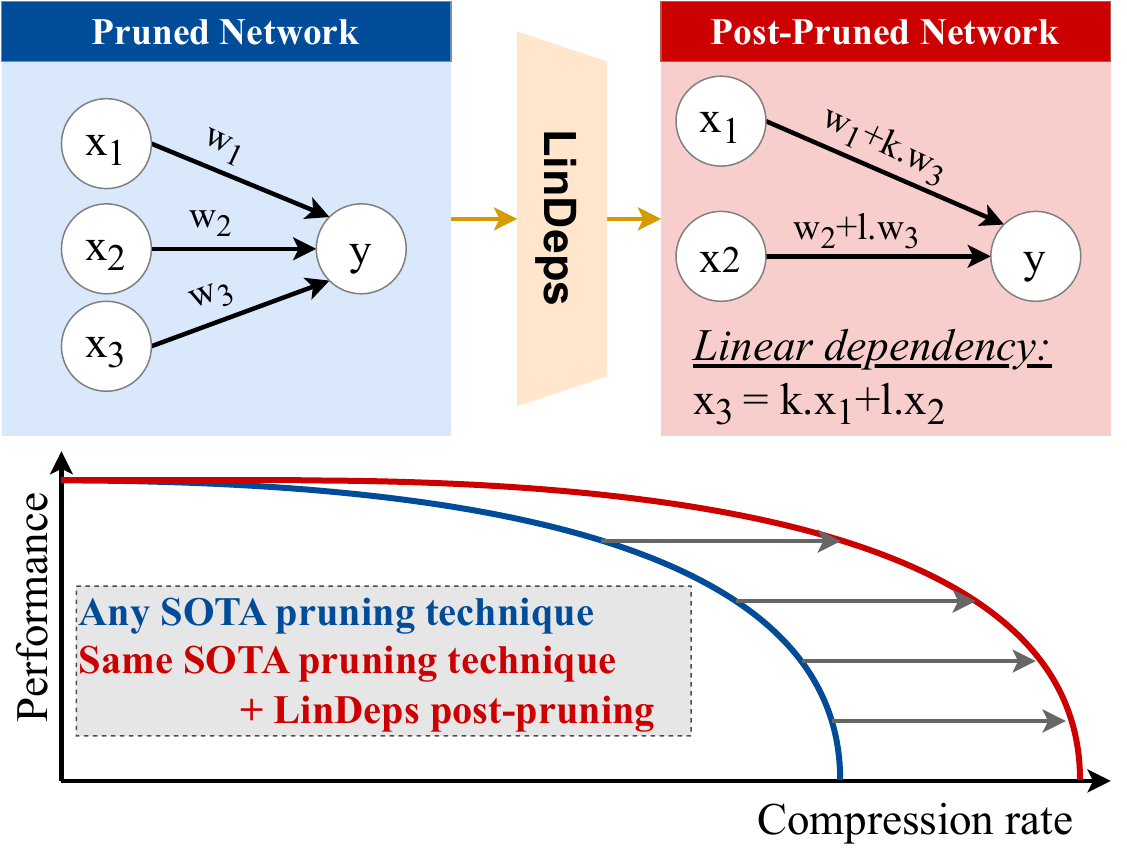}
    \caption{\textbf{Overview of our \emph{\ourMethod} post-pruning method.} We propose \ourMethod, a post-pruning method applicable on top of any pruning method. \ourMethod identifies and \emph{removes linear dependencies} between input neurons $x_i$ and output neurons $y$ of a layer using a pivoted QR decomposition, and provides a signal recovery mechanism to preserve the network’s performance \emph{without fine-tuning} leading to \emph{compression improvements with guaranteed performance preservation} of the model. Specifically, after pruning a neuron $x_i$ linearly dependent of the others, the weights $w_i$ are updated to compensate for the removal such that the same output $y$ is preserved.} 
    \label{fig:graphical_abstract}
\end{figure}

Convolutional neural networks are used in many computer vision tasks. Yet, the increasing complexity and size of state-of-the-art architectures pose significant challenges in terms of training cost, inference latency, memory footprint, and energy consumption. The need for efficient deployment of such networks on resource-constrained devices, such as mobile devices and edge computing platforms, has intensified research into network compression techniques, among which network pruning is a key solution.

Network pruning aims to reduce the size and computational requirements of neural networks by removing redundant or less important parameters while preserving as much as possible overall performance. Pruning techniques generally fall into two broad categories: unstructured pruning~\cite{Cheng2024ASurvey}, which removes individual weights, often leading to irregular sparsity that requires specialized hardware for acceleration, and structured pruning~\cite{Cheng2024ASurvey,He2024Structured}, which removes entire neurons, filters, or layers to yield smaller dense networks that benefit more from common GPU acceleration~\cite{Han2015Learning, Li2016Pruning}. While pruning has demonstrated effectiveness in reducing network size and inference latency, a fundamental challenge remains: how to optimally identify and remove redundancies without degrading network performance. 

Traditional pruning techniques rely on heuristic criteria such as weight magnitude~\cite{Han2015Learning} or norm-based importance scores~\cite{Li2016Pruning}. However, these approaches may overlook deeper structural dependencies within the network. 
Recent works have proposed alternative strategies, such as information bottleneck-based pruning~\cite{Guo2023Automatic} and redundancy-aware pruning techniques that leverage linear dependency analysis~\cite{Pan2021Network}. Nevertheless, existing techniques often fail to fully eliminate basic linear dependencies across multiple feature maps of the same network layer, leading to suboptimal pruning decisions.

In this work, we introduce \emph{\ourMethod}, a novel post-pruning method, \ie a pruning applicable on top of any pruning technique. \ourMethod systematically identifies and removes redundant filters by leveraging linear dependencies within feature maps, as illustrated in \Cref{fig:graphical_abstract}. 
In complement to traditional low-importance pruning approaches, which focus on removing the least informative neurons, \ourMethod further prunes layer-wise linear dependencies. 
By applying PQR decomposition on feature maps, our method identifies and removes linearly dependent filters. 
Importantly, we incorporate a novel signal recovery mechanism to compensate for removed parameters, thereby guaranteeing performance preservation while achieving higher compression rates, without needing to fine-tune the network. 
We experimentally demonstrate that \ourMethod significantly enhances the compression rate of state-of-the-art pruning techniques, while preserving the performance, thus establishing a new state of the art  for CNN pruning. 
Finally, we test our methods on different devices and measure the pruning time and inference speedup when the first compression method is applied without finetuning. This allows to evaluate the effectiveness of \ourMethod in setups where a finetuning would be considered too costly, such as test-time adaptation or on very scarce resources devices.

\mysection{Contributions.} 
We summarize our contributions as follows: 
\textbf{(i)} We propose \emph{\ourMethod}, a post-pruning method based on PQR decomposition to identify and remove linear dependencies in convolutional neural networks, incorporating a novel signal recovery mechanism that compensates for the removed filters, preserving the network's accuracy without requiring any additional fine-tuning, that can be combined with any other existing state-of-the-art pruning technique.
\textbf{(ii)} We demonstrate that \emph{\ourMethod} consistently improves the compression rates of existing pruning techniques while maintaining their performance, establishing new state-of-the-art pruning results.
\textbf{(iii)} In addition, we apply \emph{\ourMethod} in setups with low resources, \eg, where retraining is considered too costly or where the model needs frequent updates, and notice significant inference speedups.

\section{Related Work}\label{sec:relatedwork}

Pruning networks to increase the speed and reduce the needs in memory size and computation resources has always been tied to research on neural networks, from the early stages~\cite{Hanson1988Comparing, Lecun1989Optimal} to more recent advancements~\cite{Cheng2024ASurvey, He2024Structured, Liang2021Pruning}. 
Many variants exist, the more relevant to our work being: unstructured vs. structured pruning, data-independent vs. data-dependent pruning, low-importance vs. similarity-based pruning, and pruning of pairwise vs. layer-wise linear relations. In the following, we position \ourMethod with respect to these categories and highlight its benefits over the existing literature.

\mysection{Unstructured and structured pruning.} Network pruning can be divided into two main approaches: structured pruning that removes neurons, filters, or layers of a network, and unstructured pruning that removes individual weights. 
While unstructured pruning techniques~\cite{Frankle2018TheLottery, Guo2016Dynamic, Han2015Learning} usually allow removing more weights, they cannot be fully leveraged by most pieces of hardware. Even though some recent devices allow for acceleration of specific sparsity ratio (\eg, Hu~\etal use the NVIDIA Ampere architecture GPUs to accelerate networks with a 2:4 sparsity ratio~\cite{Hu2024Accelerating}), this solution lacks flexibility and genericity. Conversely, structured pruning~\cite{Li2017Pruning-arxiv,Liu2017Learning,He2018Soft,Lin2020HRank,He2019Filter,Pan2021Network,Guo2023Automatic,Pham2025Enhanced} does not require using specific hardware configurations and reduces the memory size needs of the network, making it a more flexible and generic solution. 
\ourMethod removes entire filters and feature maps, which thus falls within the structured pruning category and inherits its benefits over unstructured pruning.

\mysection{Data-independent and data-dependent methods.} Among structured pruning techniques, some are data independent~\cite{He2019Filter, Li2016Pruning, He2018Soft, Liu2017Learning}, meaning that they rely solely on the weights of the network for the pruning decision process, while other techniques use training data to aggregate information about the feature map distribution~\cite{Lin2020HRank, Hu2016Network-arxiv,Pan2021Network}, about gradients~\cite{VanAmersfoort2020Single-arxiv}, or retrain an auxiliary network~\cite{Ganjdanesh2024Jointly} that will make the pruning decision. 
Recent data-dependent pruning techniques, despite a slight computational overhead, tend to provide more informed pruning decisions, as they allow a better retention of task-relevant structures. 
\ourMethod follows this trend, as we use a batch of training data to find and prune linear dependencies between filters of the network.

\mysection{Low-importance and similarity pruning.} Some pruning techniques have been developed around an importance score used to rank channels within convolutional neural networks, such as HRank~\cite{Lin2020HRank}, and prune the network until a certain pruning objective is met. 
That importance score is usually used to represent how much information is contained by the associated filter so that filters carrying the least information can be pruned, yielding what is called a ``low-importance pruning''. 
A drawback of low-importance pruning techniques is that they cannot remove neurons that carry similar and significant information.

Complementary approaches, such as APIB~\cite{Guo2023Automatic} propose to use an information bottleneck principle~\cite{Tishby2015Deep}, where they evaluate the mutual information between feature maps to identify filters with similar information. 
This technique computes a score that maximizes information used for the task while reducing similarity, yielding what is called a ``similarity pruning''. 
The downside of similarity-based pruning alone is that they cannot prune neurons that convey little to no significant information within the network. 

To bridge both worlds, recent state-of-the-art techniques propose to combine low-importance and similarity-based pruning modules, such as the most complete variant of APIB~\cite{Guo2023Automatic}, resulting in hybrid approaches. 
In a different fashion, NORTON~\cite{Pham2025Enhanced}, breaks down convolutional filters into small triplets of factorized components, before pruning the triplets deemed non informative enough or too similar to another according to a custom distance function. 
We present \ourMethod as a post-pruning method, in the spirit of hybrid ones, in that we first leverage a base low-importance pruning technique, and then we apply our linear dependencies removal for maximized efficiency.

\mysection{Pruning pairwise and layer-wise linear dependencies.} APIB~\cite{Guo2023Automatic} and NORTON~\cite{Pham2025Enhanced} compute mutual information or similarity in a pairwise manner, \ie only between two filters at a time. 
This is limitative, as it cannot account for more complex linear relationships among multiple filters. 
By contrast, \ourMethod considers each filter's relationship within a layer, allowing for a comprehensive assessment of layer-wise linear dependencies and a more relevant pruning of the filters.  

Few pruning techniques have investigated structured data-dependent similarity pruning at the layer level (\ie beyond pairwise correlations). Only LDFM~\cite{Pan2021Network} goes in that direction, by leveraging a QR decomposition of a feature-based matrix and pruning the filters, yielding features linearly independent of others. 
While we use the same decomposition principle in the first step of \ourMethod, we additionally derive a novel recovery mechanism that allows us to prune the network while compensating for the pruned filters without needing fine-tuning, contrary to LDFM. 
Consequently, this allows \ourMethod to be seamlessly combined, in a post-pruning fashion, with any state-of-the-art pruning technique that focuses on identifying and removing low-importance filters, pushing pruning levels to larger values without any accuracy drop.

\mysection{Note on transformer pruning.} We developed \ourMethod on convolutional neural networks (CNNs), yet a separate and growing body of literature focuses on pruning transformer architectures~\cite{ChittyVenkata2023ASurvey,Kang2024ASurvey,Tang2024ASurvey-arxiv}. Although pruned transformers are less frequently deployed on low-resource devices due to their  larger memory and computation requirements, we will explore in future work whether \ourMethod can be effectively extended to this distinct architectural paradigm on top of existing transformer pruning techniques. 

\section{Methodology}\label{sec:method_new}

\begin{figure*}
    \centering
    \includegraphics[width=\linewidth]{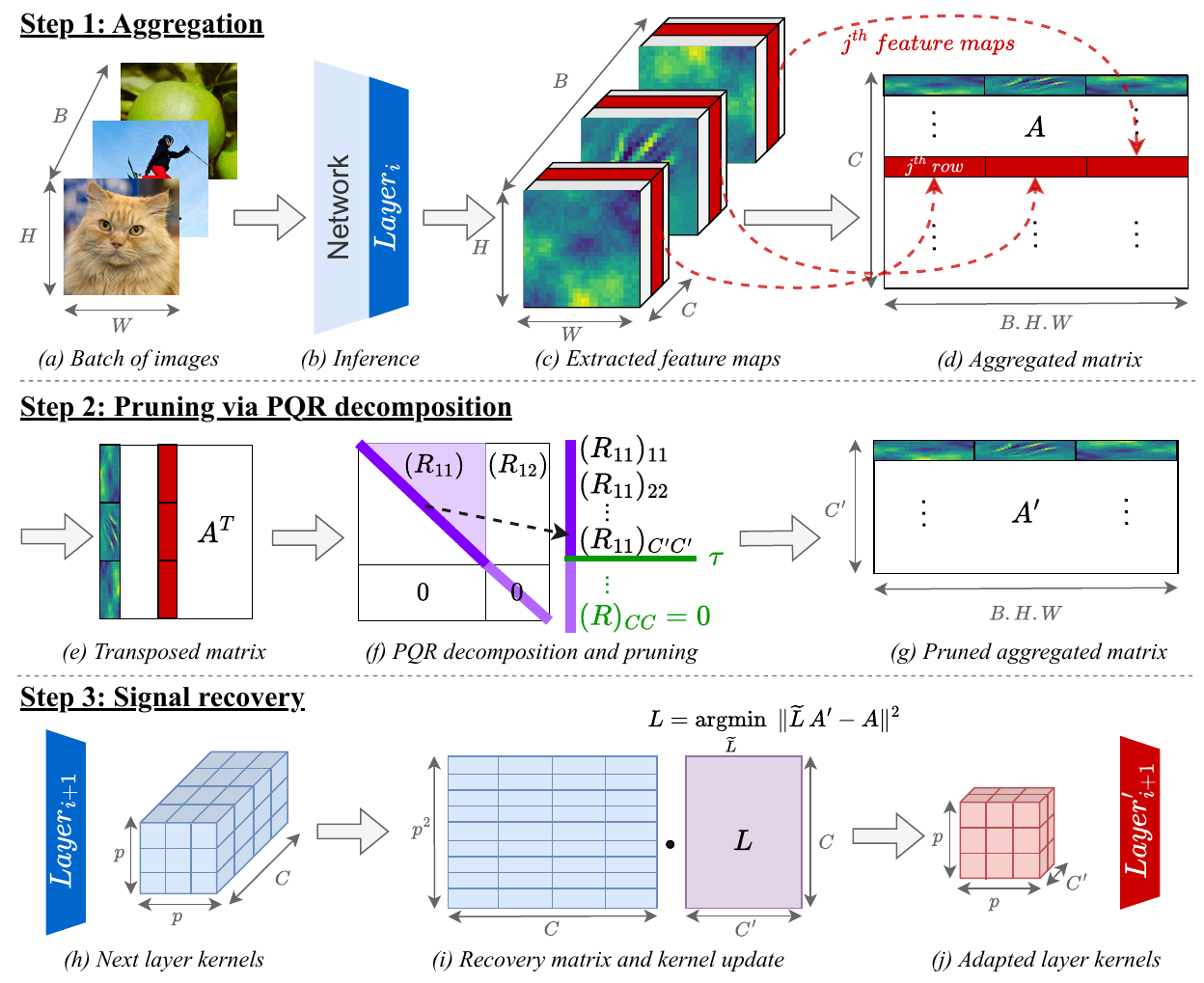}
    \caption{\textbf{Pipeline of \emph{\ourMethod} applied to a single layer.} Our method proceeds in three steps. 
    \textbf{[Step 1] Aggregation:} (a) A batch of images of size $\batch$ is passed through (b) the network up to the layer to be pruned. (c) This produces $\channels$ feature maps per image, which are (d) flattened and concatenated into a matrix $\A \in \mathbb{R}^{\channels \times (\batch \cdot \height \cdot \width)}$, where each row corresponds to one channel, aggregated over all spatial locations and all images in the batch. 
    \textbf{[Step 2]  Pruning via PQR decomposition:} (e) The aggregated matrix is transposed, and (f) a pivoted QR decomposition ($\emph{PQR}$) is applied to $\A^{\top}$, producing: $\A^{\top} \, \Pmat = \Q \R$, where $\Pmat$ permutes channels and $\R$ reveals linear dependencies on the diagonal of the upper-left matrix $\R_{11}$. 
    A threshold $\taup$ prunes (near-)dependent channels by setting to $0$ all values of $(R_{11})_{kk}<\taup ( R_{11})_{11}$, resulting in (g) a pruned matrix $\Aprime \in \mathbb{R}^{\channelsprime \times (\batch \cdot \height \cdot \width)}$, keeping only $\channelsprime$ independent feature maps. 
    \textbf{[Step 3]  Signal recovery:} (h) To adapt the next layer’s kernels, (i) a recovery matrix is computed by solving: $\Lmat \,\Aprime \approx \A$, by leveraging a least-squares error approximation. Each kernel in (h) the next layer, originally of size $\channels \times p \times p$, is (j) updated to a new kernel of size $\channelsprime \times p \times p$ by multiplying its flattened version with $\Lmat$.
}
    \label{fig:pipeline}
\end{figure*}

We first describe our pruning procedure by focusing on a single layer with \channels{} feature maps of size \height{}$\times$\width{} produced by as many convolution kernels before explaining how we sequentially prune successive layers of the network. As shown in \cref{fig:pipeline}, we proceed in three main stages, described hereafter: (1) aggregation of feature maps into a matrix, (2) pruning via \emph{PQR} decomposition, and (3) signal recovery to adapt subsequent layers. 

\mysection{Step 1: Aggregation of feature maps.} Let us define a batch size \batch{}. Passing the entire batch through the network up to the layer $i$ of interest yields \batch{} sets of feature maps, each of shape \channels{}$\times$\height{}$\times$\width{}. We \emph{flatten and concatenate} these feature maps across the batch and spatial dimensions to form a single matrix 
\(
  \A \in \mathbb{R}^{\channels \times (\batch \cdot \height \cdot \width)}.
\)
Specifically, the \(j\)-th row of \(\A\) corresponds to the flattened feature map of channel \(j\), concatenated over all \batch{} images.

\mysection{Step 2: Pruning via PQR decomposition.} Following~\cite{Pan2021Network}, to identify linearly dependent feature maps, we perform a pivoted QR decomposition (\emph{PQR})~\cite{Golub1965Numerical} 
on the transposed aggregated matrix \(\A^{\top}\). Concretely,
\begin{equation}
  \A^{\top} \, \Pmat \;=\; \Q \,\R \;=\; \Q \,\begin{bmatrix}
\R_{11} & \R_{12} \\
\mathbf{0} & \mathbf{0}
\end{bmatrix},
\end{equation}
where \(\Q\) is orthonormal, 
\(\R \in \mathbb{R}^{(\batch \cdot \height \cdot \width) \times \channels}\) with \(\R_{11} \in \mathbb{R}^{\channelsprime \times \channelsprime}\) an upper-triangular matrix, \(\channelsprime\) the effective rank of \(\A\), and \(\Pmat \in \mathbb{R}^{\channels \times \channels}\) is a permutation matrix ensuring that the diagonal elements of \(\R_{11}\) are sorted in decreasing order (in absolute value). The effective rank \(\channelsprime\) of \(\A\) measures how many rows (feature maps) are \emph{linearly independent}. 

\noindent We introduce a pruning threshold \(\taup \in [0,1)\) that determines how aggressively we discard near-dependent channels. Conceptually, we set to zero all diagonal entries of \(\mathbf{R}_{11}\) that are smaller than 
\(\taup\) times the largest diagonal element, denoted as \((\mathbf{R}_{11})_{11}\), such that:
\begin{equation}
(\mathbf{R}_{11})_{kk} \leftarrow 0, \quad \forall k \text{ where } (\mathbf{R}_{11})_{kk} < \taup (\mathbf{R}_{11})_{11}.
\end{equation}

\noindent This ensures that only the most significant independent components are retained while pruning weaker dependencies based on the threshold \(\taup\). More precisely, if \(\taup = 0\), we only prune channels that are \emph{exactly} linearly dependent; if \(\taup > 0\), we allow a mild approximation error to remove additional channels that exhibit near-dependence. In this case, we update \(\channelsprime\) to be the resulting count of channels to keep. Then, we apply the inverse of \(\Pmat\) to map back to the original indexing of \(\A\), thereby identifying \(\channelsprime\) feature maps to \emph{retain} and \(\channels - \channelsprime\) feature maps to \emph{prune}. Since each feature map is produced by a convolution kernel, we actually remove the whole kernel from the network, which incidentally prunes its corresponding feature map. 

\mysection{Step 3: Signal recovery by weight adjustment.}
To compensate for the information loss in the current layer of interest and incidentally enable pruning the next layer of the network independently of the current one, we develop a fine-tuning-free signal recovery mechanism, detailed hereafter. 
Having pruned channels at layer \(i\), we must make the subsequent layer \((i+1)\) compatible again with layer \(i\). Originally, layer \((i+1)\) expects an input of size \(\channels \times \height \times \width\). 
Now, we only provide \(\channelsprime\) channels. To ensure that the next layer’s output remains equivalent (or nearly so in the case of \(\taup>0\)) to what it was before pruning, we compute a \emph{recovery matrix} \(\Lmat\in \mathbb{R}^{\channels\times \channelsprime}\) such that
\begin{equation}
  \Lmat \,\Aprime \;\approx\; \A,
\end{equation}
where \(\Aprime \in \mathbb{R}^{\channelsprime \times (\batch \cdot \height \cdot \width)}\) is obtained by selecting the \(\channelsprime\) retained rows of \(A\), corresponding to the nonzero diagonal elements of \(\R_{11}\) after thresholding. Mathematically, each pruned channel in \(\A\) is (approximately if \(\taup>0\)) a linear combination of the retained channels in \(\Aprime\). If \(\taup=0\) and the pruned channels are strictly dependent, then \(\Lmat\) can be computed theoretically from the blocks of \(\R\). 
Nevertheless, the exact solution requires inverting \(\R_{11}\), which might be unstable numerically if its diagonal contains very small elements. 
Therefore, we rather solve for \(\Lmat\) in the least-squares sense of the Euclidean norm as follows:
\begin{equation}
  \Lmat \;=\; \underset{\widetilde{\Lmat}}{\operatorname{argmin}}\;\|\widetilde{\Lmat}\,\Aprime - \A\|^2,
\end{equation}
which also allows us to handle the case of \(\taup>0\). Each element \( (\Lmat)_{ij} \) represents the contribution of the \(j\)-th retained channel in \(\Aprime\) to reconstruct the \(i\)-th original channel in \(\A\). Finally, to accommodate the pruning performed in layer \(i\), we \emph{adapt} the convolution kernels of layer \((i+1)\) thanks to \(\Lmat\). Specifically, each kernel in layer \((i+1)\) is initially of shape \(\channels \times p\times p\) (typically $p=3$). We flatten it to a \(p^2 \times \channels\) matrix, multiply it by \(\Lmat\), and reshape it into a \(\channelsprime \times p\times p\) kernel. 
Thus, the new kernel can ingest \(\channelsprime\) channels 
instead of \(\channels\), becoming compatible with the pruned layer \(i\), and produces equivalent outputs under the no-approximation (\(\taup=0\)) scenario. Once the kernels producing feature maps of layer \((i+1)\) are updated this way, we can prune this layer \emph{independently} using the procedure described previously, ensuring that each layer’s pruning decisions do not conflict with preceding modifications. 

Eventually, to prune a whole network, we iteratively prune a layer, compensate for the information loss with our novel fine-tuning-free signal recovery mechanism, then prune the next layer, and so on. Importantly, our recovery mechanism does not need the network to be retrained or fine-tuned to compensate for a performance drop due to the pruning. 
This is a major advantage over other pruning techniques, and it allows \ourMethod to be applied efficiently after any pruning technique, providing a guaranteed extra pruning at little (\(\taup>0\)) to no (\(\taup=0\)) performance cost.

\mysection{Implementation details.} Hereafter, we provide a few generic details about our practical implementation:
\begin{itemize}
\item \emph{Batch size \(\batch\)}. We typically use \(\batch=256\), following the experimental protocols presented in~\cite{Guo2023Automatic}, although this can be adjusted if memory is constrained. Let us note that the equations governing \ourMethod hold when $\batch \cdot \height \cdot \width > \channels$, which is easily satisfied in current network architectures, even with moderate batch sizes.
\item \emph{Batch Normalization (BN)}. If a layer is immediately followed by a batch normalization module, we prune the corresponding BN parameters together with the channel.
\item \emph{Practical pruning threshold}. In practice, using $\taup=0$ retains even the smallest, numerically unstable values in the diagonal of \(\R_{11}\). Therefore, we call ``lossless pruning'' the results obtained with the application of \ourMethod with an actual value of \(\taup=10^{-6}\) instead of $\taup=0$ .
\end{itemize}

\section{Experiments}\label{sec:experiments}

We first describe the experimental settings in \cref{sec:exper_settings}, including the dataset and network architectures, the evaluation scores, the base pruning techniques, and the \ourMethod settings used.
We then present the main quantitative results in \cref{sec:mainQuantativeResults} on various datasets and network architectures, as well as a study where \ourMethod is used as a standalone pruning technique, \ie, not as a post-pruning method. Finally, in \cref{sec:low-resource}, we present results in low-resource setups, first by computing the pruning gain brought by \ourMethod over two methods when compute or time resources do not allow any retraining, second by computing the time needed to apply \ourMethod and the inference speedup gain on various devices, from consumer laptops to edge devices.

\subsection{Experimental settings}
\label{sec:exper_settings}

\mysection{Datasets and network architectures.}
We assess the pruning capabilities of our post-pruning \ourMethod method in combination with existing pruning techniques on two datasets: CIFAR-10~\cite{Krizhevsky2009Learning} and ImageNet~\cite{Deng2009ImageNet}. On CIFAR-10, we compare three architectures: VGG-16~\cite{Simonyan2014VeryDeep-arxiv}, ResNet-56~\cite{He2016DeepResidual}, and ResNet-110~\cite{He2016DeepResidual}. On ImageNet, we use the ResNet-50 architecture, following common practice in pruning benchmarks~\cite{Pham2025Enhanced}.

\mysection{Evaluation scores.}
We assess the compression induced by \ourMethod in comparison to existing pruning techniques using the reduction ratio of
FLOPs (floating-point operations) and the reduction ratio of the number of network parameters. 
We measure the amount of FLOPs and parameters using the \emph{ptflop}~\cite{Sovrasov2024ptflops} library.
To measure the pruned network performance, we report the corresponding top-1 accuracy, in line with standard benchmarks commonly used in the pruning literature. Furthermore, it is the most neutral choice of ranking score in terms of application preferences, as demonstrated in Pi\'erard~\etal~\cite{Pierard2025Foundations}.

\mysection{Base pruning techniques.} 
We study the effect of our proposed post-pruning \ourMethod method on top of two of the latest state-of-the-art pruning techniques: APIB~\cite{Guo2023Automatic} released in 2023, and NORTON~\cite{Pham2025Enhanced} published in 2025. For completeness, we also provide the results of \ourMethod without a base pruning technique.
To combine the APIB pruning technique with \ourMethod, we first reproduced the results of the pruned network by using the code released by the authors on GitHub\footnote{\url{https://github.com/sunggo/APIB}}. 
After several trainings with different seeds, we selected the baseline VGG-16 as the one whose accuracy before pruning is the closest to the one reported in the original paper~\cite{Guo2023Automatic}. We kept all training and post-pruning fine-tuning settings similar to the ones reported by the authors, \ie, all networks are trained using the Stochastic Gradient Descent (SGD) optimizer with a momentum of $0.9$ and a weight decay of $2 \times 10^{-4}$. 
The initial learning rate is set to $0.1$ and decayed using the cosine annealing scheduling~\cite{Loshchilov2017SGDR}. The batch size is set to $256$ and the number of epochs to $350$.
As APIB allows targeting any amount of pruning, we pruned several networks with pruning ratios between $65\%$ and $80\%$.
For each amount of pruning, we fine-tuned the network obtained as described in~\cite{Guo2023Automatic}. To avoid the seed cherry-picking reproducibility issue, we repeated the fine-tuning process eight times from the start and reported the average accuracy over the eight trials. Regarding \ourMethod, the eight fine-tuned networks were post-pruned without additional fine-tuning, and the performances in terms of top-1 accuracies, FLOPs, and parameters were averaged, also to ensure reproducibility. 
Similarly, we combined the current state-of-the-art NORTON pruning technique~\cite{Pham2025Enhanced} with \ourMethod. To do so, we also first reproduced the results of the paper through the code provided on the authors' GitHub\footnote{\url{https://github.com/pvti/NORTON}}. As a base, we used the checkpoints of the pruned and retrained networks available from the GitHub page, without retraining the networks ourselves. We then applied \ourMethod without additional fine-tuning and reported the accuracy, FLOPs, and parameters.

\subsection{Main quantitative results}\label{sec:mainQuantativeResults}
In this section, we present the main quantitative results of our post-pruning \ourMethod method in the performance-preserving setting applied on top of APIB and NORTON on various datasets and with different network architectures.

\begin{figure}[t]
    \centering
    \includegraphics[width=\linewidth]{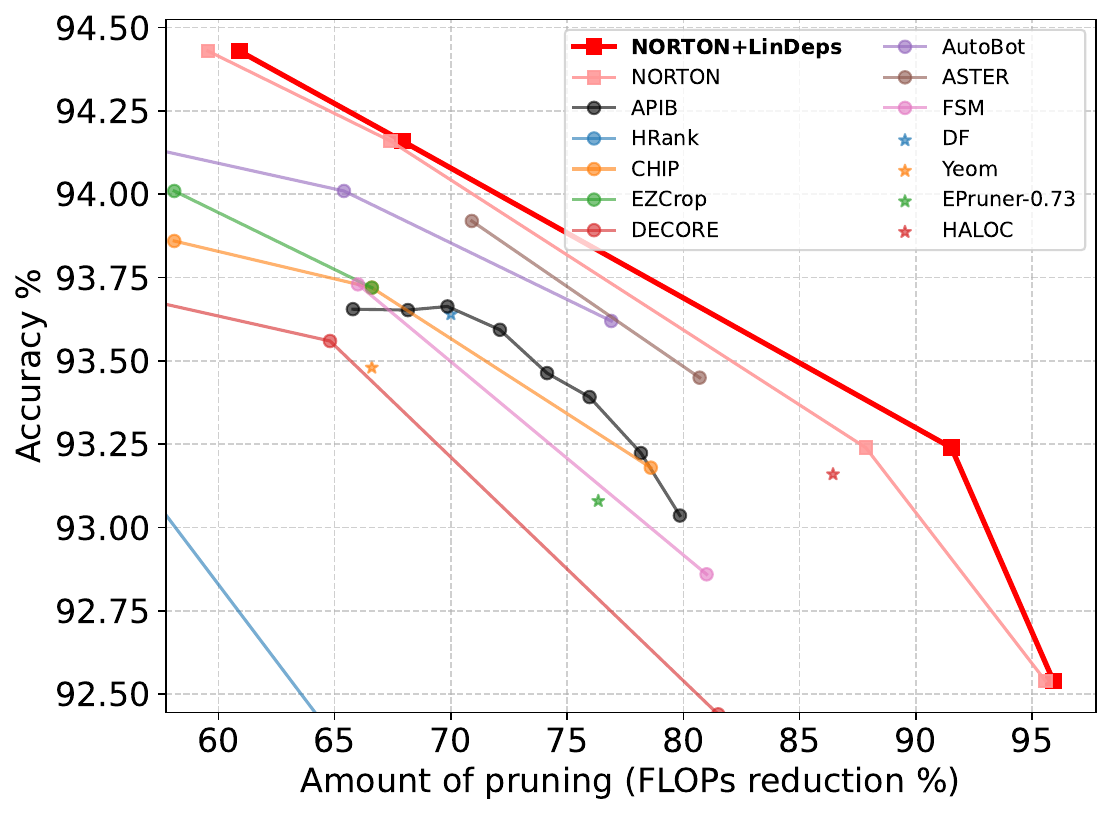}
    \caption{ \textbf{Benchmark of pruning techniques with VGG-16 trained on CIFAR-10.} \ourMethod, applied on top of NORTON, achieves a new state of the art, surpassing all previously published pruning techniques by a significant margin in terms of compression level while maintaining the top-1 accuracy.}
    \label{fig:Benchmark_SOTA_VGG16_CIFAR10}
\end{figure}

\mysection{VGG-16 on CIFAR-10.} The results of post-pruning using \ourMethod in combination with APIB and NORTON on CIFAR-10 are reported in~\cref{tab:pruning_results_VGG16_CIFAR-10}. 
For all compression levels, \ourMethod increases the reduction ratios while keeping the original top-1 accuracy. Particularly, with APIB, \ourMethod increases the number of FLOPs pruned by an additional 1.23\%, 1.26\%, 1.25\%, 1.01\%, 0.91\%, 0.96\%, 0.71\%, and 0.54\% for initial reduction ratios ranging from 65.78\% to 79.85\%. Combined with NORTON, \ourMethod increases the reduction ratios by 1.39\%, 0.52\%, 5.7\%, and 0.54\% for initial reduction ratios ranging between 59.58\% and 95.58\%. 
Furthermore, we provide a complete benchmark of pruning techniques in~\cref{fig:Benchmark_SOTA_VGG16_CIFAR10}. As illustrated, \ourMethod combined with NORTON leads to a new state-of-the-art performance, outperforming previously published techniques. 

\mysection{ResNet-56/110 on CIFAR-10.} \Cref{tab:pruning_results_Resnet56_CIFAR10,tab:pruning_results_Resnet110_CIFAR-10} respectively, show the results of \ourMethod combined with NORTON with a ResNet-56 and ResNet-110 network architecture on CIFAR-10.  In the performance-preserving regime, \ourMethod similarly pushes the pruning ratio of FLOPs by an additional 1.44\%, 0.16\%, 2.57\%, and 1.1\% on ResNet-56 for reduction ratios ranging from 26.69\% to 89.20\% and 2.82\%, 0.48\%, and 0.21\% on ResNet-110 for ratios ranging from 36.87\% to 81.26\%.
This shows that \ourMethod works on both small and large network architectures.

\mysection{ResNet-50 on ImageNet.} The results for \ourMethod applied on top of NORTON with a ResNet-50 architecture trained on ImageNet are presented in~\cref{tab:pruning_results_Resnet50_ImageNet}. We still observe an improvement for each compression level, up to $0.17\%$ on the pruning ratio while keeping the same top-1 accuracy. Hence, this shows that even when trained on large datasets, our post-pruning \ourMethod method still allows pushing the compression level at no performance cost. In this case, the improvement is smaller than the one obtained on CIFAR-10. This could be because ResNet-50 under-performs too much on the ImageNet dataset (only around $75\%$ accuracy), and thus the feature maps may contain a lot of noise, making it harder to find linear relationships.

\begin{table}[ht]
    \centering
    \resizebox{\columnwidth}{!}{%
    \begin{tabular}{l p{2cm} p{1.8cm} p{1.5cm} p{1.5cm}}
        \toprule
        \textbf{Pruning techniques} & \textbf{Baseline Top-1~Acc.} & \textbf{Pruned Top-1~Acc.} & \textbf{FLOPs reduc. ↑} & \textbf{Params. reduc. ↑} \\

        \midrule
        HRank~\cite{Lin2020HRank}& 93.96\% & 93.43\% & 53.50\% & 82.90\% \\
        CHIP~\cite{Sui2021CHIP} & 93.96\% & 93.86\% & 58.10\% & 81.60\% \\
        EZCrop~\cite{Lin2022EZCrop}& 93.96\% & 94.01\% & 58.10\% & 81.60\% \\
        DECORE-500~\cite{Alwani2022DECORE}& 93.96\% & 94.02\% & 35.30\% & 63.00\% \\
        AutoBot~\cite{Castells2021Automatic-arxiv} & 93.96\% & 94.19\% & 53.70\% & 49.80\% \\

        APIB~\cite{Guo2023Automatic} & 93.68\% & 94.08\% & 60.00\% & 76.00\% \\
        NORTON~\cite{Pham2025Enhanced} $\triangle$ & 93.96\% & 94.43\% & 59.58\% & 82.72\% \\
        \colorize{NORTON+\ourMethod (ours)} & 93.96\% & 94.43\% & 60.97\% & 83.02\% \\
        
        \midrule
        HRank~\cite{Lin2020HRank} & 93.96\% & 92.34\% & 65.30\% & 82.10\% \\
        CHIP~\cite{Sui2021CHIP} & 93.96\% & 93.72\% & 66.60\% & 83.30\% \\
        EZCrop~\cite{Lin2022EZCrop} & 93.96\% & 93.72\% & 66.60\% & 83.30\% \\
        DECORE-200~\cite{Alwani2022DECORE} & 93.96\% & 93.56\% & 64.80\% & 89.00\% \\
        AutoBot~\cite{Castells2021Automatic-arxiv} & 93.96\% & 94.01\% & 65.40\% & 57.00\% \\
        Yeom~\etal~\cite{Yeom2021Toward-arxiv} & 93.96\% & 93.48\% & 66.60\% & 80.90\% \\

        DF~\cite{Eo2023ADifferentiable-arxiv} & 94.14\% & 93.64\% & 70.00\% & 90.10\% \\
        FSM~\cite{Duan2023Network} & 93.96\% & 93.73\% & 66.00\% & 86.30\% \\

        APIB~\cite{Guo2023Automatic} & 93.68\% & 94.00\% & 66.00\% & 78.00\% \\
        APIB $*$ & 93.69\% & 93.65\% & 65.78\% & 89.59\% \\
        \colorize{APIB+\ourMethod (ours)} & {93.69\%} & {93.65\%} & {67.01\%} & {90.17\% }\\
        NORTON~\cite{Pham2025Enhanced} $\triangle$ & 93.96\% & 94.16\% & 67.43\% & 84.30\% \\
        \colorize{NORTON+\ourMethod (ours)} & 93.96\% & 94.16\% & 67.95\% & 84.38\% \\
        
        \midrule
        CHIP~\cite{Sui2021CHIP} & 93.96\% & 93.18\% & 78.60\% & 87.30\% \\
        DECORE-100~\cite{Alwani2022DECORE} & 93.96\% & 92.44\% & 81.50\% & 96.60\% \\
        FSM~\cite{Duan2023Network} & 93.96\% & 92.86\% & 81.00\% & 90.60\% \\
        EPruner-0.73~\cite{Lin2022Network} & 93.02\% & 93.08\% & 76.34\% & 88.80\% \\
        
        HALOC~\cite{Xiao2023HALOC} & 92.78\% & 93.16\% & 86.44\% & 98.56\% \\
        
        ASTER~\cite{Zhang2024Adaptive} & 93.90\% & 93.92\% & 70.90\% & N/A \\
        ASTER~\cite{Zhang2024Adaptive} & 92.90\% & 93.45\% & 80.70\% & N/A \\
        
        AutoBot~\cite{Castells2021Automatic-arxiv} & 93.96\% & 93.62\% & 76.90\% & 63.24\% \\
        
        APIB $*$ & 93.69\% & 93.65\% & 68.15\% & 91.00\% \\
        \colorize{APIB+\ourMethod (ours)}& 93.69\% & 93.65\% & 69.41\% & 91.49\% \\

        APIB $*$ & 93.69\% & 93.66\% & 69.85\% & 92.00\% \\
        \colorize{APIB+\ourMethod (ours)}& 93.69\% & 93.66\% & 71.10\% & 92.41\% \\
        
        APIB $*$ & 93.69\% & 93.59\% & 72.10\% &  92.58\%\\
        \colorize{APIB+\ourMethod (ours)}& 93.69\% & 93.59\% & 73.11\% & 92.99\% \\
        
        APIB $*$ & 93.69\% & 93.46\% & 74.13\% &  93.39\%\\
        \colorize{APIB+\ourMethod (ours)}& 93.69\% & 93.46\% & 75.04\% & 93.72\% \\

        APIB $*$ & 93.69\% & 93.39\% & 75.96\% &  94.29\%\\
        \colorize{APIB+\ourMethod (ours)}& 93.69\% & 93.39\% & 76.92\% & 94.61\% \\

        APIB $*$ & 93.69\% & 93.22\% & 78.17\% & 95.14\% \\
       \colorize{APIB+\ourMethod (ours)}& 93.69\% & 93.22\% & 78.88\% & 95.42\% \\

        APIB $*$ & 93.69\% & 93.03\% & 79.85\% & 95.80\% \\
        \colorize{APIB+\ourMethod (ours)}& 93.69\% & 93.03\% & 80.39\% & 96.07\% \\
        
        NORTON~\cite{Pham2025Enhanced} $\triangle$ & 93.96\% & 93.24\% & 87.86\% & 87.04\% \\
        \colorize{NORTON+\ourMethod (ours)} & 93.96\% & 93.24\% & 93.56\% & 96.28\% \\

        \midrule
        HRank~\cite{Lin2020HRank} & 93.96\% & 91.23\% & 76.50\% & 92.00\% \\
        RGP~\cite{Chen2024RGP} & 93.14\% & 91.45\% & 90.49\% & 90.62\% \\
        DECORE-50~\cite{Alwani2022DECORE} & 93.96\% & 91.68\% & 88.30\% & 98.30\% \\

        NORTON~\cite{Pham2025Enhanced} $\triangle$& 93.96\% & 92.54\% & 95.58\% & 98.33\% \\
        \colorize{NORTON+\ourMethod (ours)} & 93.96\% & {92.54\%} & {96.12\%} & {98.66\%} \\
    \end{tabular}
    }
    \caption{\textbf{Performance comparison of pruning techniques for VGG-16 on CIFAR-10} roughly grouped by FLOPs reduction as in~\cite{Pham2025Enhanced}. The following symbols apply: no symbol means that the results are directly imported from the original paper, $*$ refers to results reproduced using the code from the official GitHub, and $\triangle$ means that the results are obtained from the available code and checkpoints from the official GitHub. Applying \ourMethod consistently increases FLOPs reduction and parameters reduction.}
    \label{tab:pruning_results_VGG16_CIFAR-10}
\end{table}

\begin{table}[ht]
    \centering
    \resizebox{\columnwidth}{!}{%
    \begin{tabular}{l p{2cm} p{1.8cm} p{1.5cm} p{1.5cm}}
        \toprule
        \textbf{Pruning techniques} & \textbf{Baseline Top-1 Acc.} & \textbf{Pruned Top-1 Acc.} & \textbf{FLOPs reduc. ↑} & \textbf{Params. reduc. ↑} \\
        \midrule

        HRank~\cite{Lin2020HRank} & 93.26\% & 93.52\% & 29.30\% & 16.80\% \\
        DECORE-450~\cite{Alwani2022DECORE} & 93.26\% & 93.34\% & 26.30\% & 24.20\% \\
        
        NORTON~\cite{Pham2025Enhanced} $\triangle$& 93.26\% & 94.48\% & 25.25\% & 30.77\% \\
       \colorize{NORTON+\ourMethod (ours)} & 93.26\% & 94.48\% & 26.69\% & 31.06\% \\

        \midrule
        HRank~\cite{Lin2020HRank} & 93.26\% & 93.17\% & 50.00\% & 42.4\% \\
        DECORE-200~\cite{Alwani2022DECORE} & 93.26\% & 93.26\% & 49.90\% & 49.00\% \\
        FSM~\cite{Duan2023Network} & 93.26\% & 93.63\% & 51.20\% & 43.60\% \\
        AutoBot~\cite{Castells2021Automatic-arxiv} & 93.27\% & 93.76\% & 55.9\% & 45.9\% \\
        EZCrop~\cite{Lin2022EZCrop} & 93.26\% & 93.80\% & 47.40\% & 42.80\% \\
        
        NORTON~\cite{Pham2025Enhanced} $\triangle$& 93.26\% & 93.99\% & 41.33\% & 47.31\% \\
        \colorize{NORTON+\ourMethod (ours)} & 93.26\% & 93.99\% & 41.49\% & 47.34\% \\
        
        \midrule
        CHIP~\cite{Sui2021CHIP} & 93.26\% & 92.05\% & 71.80\% & 72.30\% \\
        APIB~\cite{Guo2023Automatic} & 93.26\% & 93.29\% & 67.00\% & 66.00\% \\
        
        NORTON~\cite{Pham2025Enhanced} $\triangle$& 93.26\% & 93.81\% & 69.67\% & 74.39\% \\
        \colorize{NORTON+\ourMethod (ours)} & 93.26\% & 93.81\% & 72.24\% & 76.72\% \\

        \midrule
        DECORE-55~\cite{Alwani2022DECORE} & 93.26\% & 90.85\% & 81.50\% & 85.30\% \\
        APIB~\cite{Guo2023Automatic} & 93.26\% & 91.53\% & 81.00\% & 83.00\% \\

        NORTON~\cite{Pham2025Enhanced} $\triangle$& 93.26\% & 91.61\% & 88.10\% & 89.69\% \\
        \colorize{NORTON+\ourMethod (ours)} & 93.26\% & 91.61\% & 89.20\% & 91.23\%\\
    \end{tabular}%
    }
    \caption{\textbf{Pruning results with ResNet-56 trained on CIFAR-10}, comparing NORTON with and without \ourMethod. In the performance-preserving regime, \ourMethod consistently increases the pruning ratio of FLOPs across all compression levels, with gains ranging from $0.16\%$ to $2.57\%$. The symbol conventions from  \cref{tab:pruning_results_VGG16_CIFAR-10} apply.}
    \label{tab:pruning_results_Resnet56_CIFAR10}
\end{table}

\begin{table}[ht]
    \centering
    \resizebox{\columnwidth}{!}{%
    \begin{tabular}{l p{2cm} p{1.8cm} p{1.5cm} p{1.5cm}}
        \toprule
        \textbf{Pruning techniques} & \textbf{Baseline Top-1 Acc.} & \textbf{Pruned Top-1 Acc.} & \textbf{FLOPs reduc. ↑} & \textbf{Params. reduc. ↑} \\
        \midrule
        DECORE-500~\cite{Alwani2022DECORE} & 93.50\% & 93.88\% & 35.40\% & 35.70\% \\
        
        NORTON~\cite{Pham2025Enhanced} $\triangle$& 93.50\% & 94.85\% & 34.05\% & 37.16\% \\
        \colorize{NORTON+\ourMethod (ours)} & 93.50\% & 94.85\% & 36.87\% & 37.72\% \\

        \midrule
        DECORE-300~\cite{Alwani2022DECORE} & 93.50\% & 93.50\% & 61.80\% & 64.80\% \\
        
        NORTON~\cite{Pham2025Enhanced} $\triangle$& 93.50\% & 94.05\% & 63.00\% & 65.13\% \\
        \colorize{NORTON+\ourMethod (ours)} & 93.50\% & 94.05\% & 63.48\% & 65.22\%\\

        \midrule
        DECORE-175~\cite{Alwani2022DECORE} & 93.50\% & 92.71\% & 76.90\% & 79.60\% \\

        NORTON~\cite{Pham2025Enhanced} $\triangle$& 93.50\% & 92.77\% & 81.05\% & 82.41\% \\
        \colorize{NORTON+\ourMethod (ours)} & 93.50\% & 92.77\% & 81.26\% & 82.48\% \\

    \end{tabular}%
    }
    \caption{\textbf{Pruning results with ResNet-110 trained on CIFAR-10}, comparing NORTON with and without \ourMethod. Similarly to ResNet-56, \ourMethod enhances the pruning ratio of FLOPs across all compression levels, with improvements ranging from $0.21\%$ to $2.82\%$. The symbol conventions from  \cref{tab:pruning_results_VGG16_CIFAR-10} apply.}
    \label{tab:pruning_results_Resnet110_CIFAR-10}
\end{table}

\begin{table}[ht]
    \centering
    \resizebox{\columnwidth}{!}{%
    \begin{tabular}{l p{2cm} p{1.8cm} p{1.5cm} p{1.5cm}}
        \toprule
        \textbf{Pruning techniques} & \textbf{Baseline Top-1 Acc.} & \textbf{Pruned Top-1 Acc.} & \textbf{FLOPs reduc. ↑} & \textbf{Params. reduc. ↑} \\
        \midrule
        
        NORTON~\cite{Pham2025Enhanced} $\triangle$& 76.15\% & 76.90\% & 42.91\% & 43.06\% \\
        \colorize{NORTON+\ourMethod (ours)} & 76.15\% & 76.90\% & 43.08\% & 43.10\% \\

        \midrule
        NORTON~\cite{Pham2025Enhanced} $\triangle$& 76.15\% & 76.59\% & 48.82\% & 47.03\% \\
        \colorize{NORTON+\ourMethod (ours)} & 76.15\% & 76.59\% & 48.85\% & 47.05\% \\

        \midrule
        NORTON~\cite{Pham2025Enhanced} $\triangle$& 76.15\% & 75.95\% & 63.16\% & 58.72\% \\
        \colorize{NORTON+\ourMethod (ours)} & 76.15\% & 75.95\% & 63.24\% & 58.73\% \\

        \midrule
        NORTON~\cite{Pham2025Enhanced} $\triangle$& 76.15\% & 74.00\% & 75.80\% & 68.78\% \\
        \colorize{NORTON+\ourMethod (ours)} & 76.15\% & 74.00\% & 75.85\% & 68.79\% \\
        
        \midrule
        NORTON~\cite{Pham2025Enhanced} $\triangle$& 76.15\% & 73.65\% & 77.22\% & 76.91\% \\
        \colorize{NORTON+\ourMethod (ours)} & 76.15\% & 73.65\% & 77.33\% & 76.93\% \\

    \end{tabular}%
    }
    \caption{\textbf{Pruning results with ResNet-50 trained on ImageNet}, comparing NORTON with and without \ourMethod. Similarly, \ourMethod further improves the pruning ratio by up to $0.17\%$ at identical top-1 accuracy. The symbol conventions from Table \ref{tab:pruning_results_VGG16_CIFAR-10} apply.}
    \label{tab:pruning_results_Resnet50_ImageNet}
\end{table}

\mysection{\ourMethod as a standalone pruning technique.}
We also evaluated \ourMethod as a standalone pruning technique, applied directly to VGG-16 trained on CIFAR-10, without relying on any base pruning technique. The results, reported for different threshold values $\taup$, show that \ourMethod alone achieves the following pruning ratios and corresponding top-1 accuracies: for $\taup = 0$, \ourMethod reaches a pruning ratio of 24.9\% with no accuracy loss (93.68\% top-1 accuracy), demonstrating that nearly 25\% of the parameters can be safely removed at no performance cost by eliminating strictly linearly dependent filters. However, as the pruning threshold $\taup$ increases between $0.05$ and $0.5$, the pruning ratios increase to 34.9\%, 40.8\%, 43.2\%, and 64.2\%, but at the cost of significant performance degradation, with top-1 accuracies dropping to 92.65\%, 91.08\%, 78.60\%, and 23.2\%, respectively. 
This confirms that \ourMethod alone does not achieve competitive performance compared to modern pruning techniques (see~\cref{tab:pruning_results_VGG16_CIFAR-10}), as it was designed to complement existing pruning techniques rather than replace them. The lack of re-fine-tuning after pruning further exacerbates this drop, as no training step compensates for the removed filters. 
Additionally, the purely linear dependency-based approach of \ourMethod only captures correlations between highly similar feature maps, but fails to target low-importance neurons that could be pruned independently regarding other neurons, which are typically removed by conventional importance-based pruning techniques. 
This highlights why \ourMethod performs best when combined with other pruning techniques, leveraging complementary pruning criteria to maximize both compression and performance preservation.

\subsection{\ourMethod in a low-resource setup}\label{sec:low-resource}
In this section, we consider the practical case of low-resource environments where a model can be pruned but the resources to retrain it are not available. This can be the case when the retraining process is long and/or computationally expensive, or when regular and fast updates need to be made on the model, \eg, in the case of test-time adaptation~\cite{Cioppa2019ARTHuS}. 

\begin{figure}[t]
    \centering
    \includegraphics[width=\linewidth]{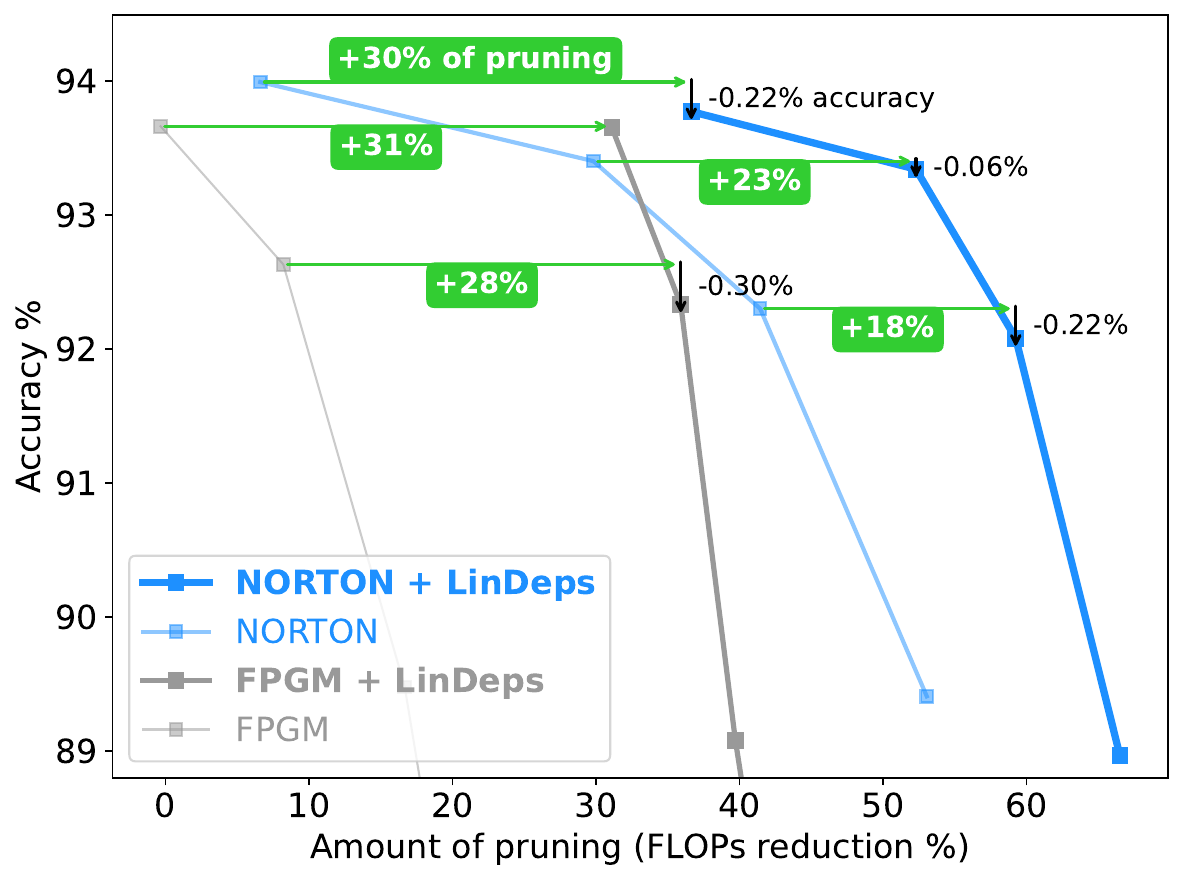}
    \caption{ \textbf{\ourMethod pruning improvement without retraining.} In low-resource environments where post-pruning retraining cannot be afforded, \ourMethod gives a significant pruning boost for a comparatively negligible decrease in performance.}
    \label{fig:method+lindeps}
\end{figure}

In that regard, \cref{fig:method+lindeps} shows the pruning gain obtained by applying \ourMethod  on top of a VGG model trained on CIFAR-10 then compressed with NORTON (using their low-rank method with a rank of $6$), but not retrained after pruning. We used $\taup=0.1$ which empirically gave the best trade-off between pruning gain and accuracy loss when pruning without retraining. We achieve gains of up to $30\%$ of pruning with losses in accuracy lower than $0.3\%$, highlighting the benefits of \ourMethod in this setup.  

Furthermore, \cref{tab:device_times} compares the time needed to perform NORTON and \ourMethod on several devices, from consumer laptops to resource-limited systems, as well as the gain in inference time for a batch of $256$ images before and after applying \ourMethod (after the NORTON pruning). We can observe that, comparatively, running \ourMethod takes significantly less time than running NORTON, and that \ourMethod provides a consistent and valuable speedup in inference, which is critical in real-time applications\footnote{We also note that some operations in NORTON were not supported by the MPS chip of the MacBook, defaulting to the CPU whenever needed, which might explain the large corresponding time in \cref{tab:device_times}}. 

\begin{table}[!ht]
\centering
\resizebox{\columnwidth}{!}{%
\begin{tabular}{p{3.3cm} p{1.5cm} p{1.5cm} p{1.5cm}}
\toprule
Device & {NORTON time (s)} & {\ourMethod time (s)} & {Inference latency} \\
\midrule
Asus GPU RTX 3070      & 607  & 13 & -23\% \\
Asus CPU Intel Core i7    & 460 & 33 & -21\% \\
MacBook M1 MPS      & 8,872  & 30 & -29\% \\
MacBook M1 CPU      & 288 & 111 & -33\% \\
Orin AGX GPU      & 3,718 & 17 & -25\% \\
Orin AGX CPU      & 1,903 & 403 & -23\% \\
Raspberry Pi CPU      & 1,738 & 247 & -28\% \\
\bottomrule
\end{tabular}
}
\caption{\textbf{Time benchmark of \ourMethod.} Comparison of NORTON and \ourMethod runtimes across consumer laptops (Asus laptop with a NVIDIA GeForce RTX 3070, MacBook Pro M1) and low-resource devices (Orin AGX, Raspberry Pi 5), and inference latency gain by applying \ourMethod on batches of 256 images.}
\label{tab:device_times}
\end{table}

\section{Conclusion}\label{sec:conclusion}

In this paper, we introduced \emph{\ourMethod}, a novel post-pruning method designed to systematically identify and eliminate redundant filters in convolutional networks through the analysis of linear dependencies within feature maps. By leveraging pivoted QR decomposition, \ourMethod effectively detects dependent filters and removes them, while a novel signal recovery mechanism adjusts the subsequent layer to preserve the network's compatibility and performance.
A key advantage of \ourMethod is its flexibility: it can be applied on top of any existing pruning technique, boosting its compression rate with either zero or controlled accuracy degradation, depending on the desired compression rate. This versatility makes it a valuable complement to state-of-the-art pruning techniques across various architectures and datasets.
Our experiments demonstrated that \ourMethod consistently improves the compression rates of multiple pruning baselines on CIFAR-10 and ImageNet, achieving new state-of-the-art results. Furthermore, we highlighted that \ourMethod performs well on both small and large networks.
Overall, \ourMethod is an efficient and theoretically grounded approach to post-pruning, contributing to the broader goal of enabling lightweight yet high-performing deep learning networks for low-resource devices.

\noindent\textbf{Acknowledgments.}
The research of M.~Henry was funded by the PIT MecaTech under grant No.~C8650 (ReconnAIssance). 
The research of A.~Deli{\`e}ge was funded by the \href{https://www.frs-fnrs.be}{F.R.S.-FNRS} under project grant T$.0065.22$ and by the SPW EER, Wallonia, Belgium, under grant n°2010235 (ARIAC by \href{https://www.digitalwallonia.be/en/}{DIGITALWALLONIA4.AI}).

\end{document}